\documentclass[runningheads]{llncs}
\usepackage[T1]{fontenc}
\usepackage[]{graphicx}
\usepackage{subcaption}
\usepackage{amsmath,amssymb,amsfonts}
\usepackage{gensymb}
\usepackage{siunitx}
\usepackage{cite}
\usepackage{graphics,multirow} 

\usepackage{csquotes}
\usepackage{placeins}
\usepackage{xcolor}

\usepackage{textcomp}
\usepackage{siunitx}
\usepackage{bm}
\usepackage{booktabs}
\usepackage{threeparttable}
\usepackage[misc]{ifsym}

\begin{document}

\title{SLAM-TKA: Real-time Intra-operative Measurement of Tibial Resection Plane in Conventional Total Knee Arthroplasty%
\thanks{This work was supported in part by the Australian Research Council Discovery Project (No. DP200100982), and in part by Shenzhen Fundamental Research Program (No: JCYJ20200109141622964).}}
\titlerunning{SLAM-TKA}
%

\author{Shuai Zhang\inst{1,2} \and
	Liang Zhao\inst{1}\textsuperscript{(\Letter)}  \and
	Shoudong Huang\inst{1} \and 
	Hua Wang\inst{3}\textsuperscript{(\Letter)}  \and 
	Qi Luo\inst{3} \and 
	Qi Hao\inst{2}}


\authorrunning{S. Zhang et al.}

\institute{Robotics Institute, University of Technology Sydney, Sydney, Australia \email{Liang.Zhao@uts.edu.au}
	\and
	Department of Computer Science and Engineering, Southern University of Science and Technology, Shenzhen, 518055, China \and
	Osteoarthropathy Surgery Department, Shenzhen People's Hospital,  Shenzhen, 518020, China
}

\maketitle              
\begin{abstract}
Total knee arthroplasty (TKA) is a common orthopaedic surgery to replace a damaged knee joint with artificial implants. The inaccuracy of achieving the planned implant position can result in the risk of implant component aseptic loosening, wear out, and even a joint revision, and those failures most of the time occur on the tibial side in the conventional jig-based TKA (CON-TKA). This study aims to precisely evaluate the accuracy of the proximal tibial resection plane intra-operatively in real-time such that the evaluation processing changes very little on the CON-TKA operative procedure.
Two X-ray radiographs captured during the proximal tibial resection phase together with a pre-operative patient-specific tibia 3D mesh model segmented from computed tomography (CT) scans and a trocar pin 3D mesh model are used in the proposed simultaneous localisation and mapping (SLAM) system to estimate the proximal tibial resection plane.
Validations using both simulation and in-vivo datasets are performed to demonstrate the robustness and the potential clinical value of the proposed algorithm.

\end{abstract}

\section{Introduction}
\label{sec:introduction}
Total knee arthroplasty (TKA) is considered to be the gold standard to relieve disability and to reduce pain for end-stage knee osteoarthritis. It is the surgery to replace a knee joint with an artificial joint. The number of TKA surgeries has increased at a dramatic rate throughout the world due to the growing prevalence of knee arthritis and the increased access to orthopaedic care.
For example, the number of TKA procedures in USA is projected to grow to 1.26 million by 2030 \cite{gao2020primary}. Meanwhile, the frequency of TKA revisions has also increased and the most common reason is aseptic loosening caused by implant malalignment \cite{pietrzak2019have}.

Regarding the optimal implant alignment, studies have confirmed that the distal femoral and proximal tibial should be resected perpendicular to their mechanical axes \cite{gromov2014optimal} (the standard accuracy requirement is less than \ang{3} angle error).
Otherwise, the inaccuracy of the distal femoral and proximal tibial resection will result in an increased risk of component aseptic loosening and early TKA failure, and those failures often occur on the tibial side \cite{berend2004chetranjan}.
However, as the most commonly used alignment guide device in the conventional jig-based TKA (CON-TKA), the extramedullary (EM) alignment guide for the proximal tibial resection shows a limited degree of accuracy \cite{iorio2013accuracy}.
Studies have shown that roughly 12.4\% and 36\% of patients who underwent CON-TKA procedures have the coronal and saggital tibial component angle errors more than \ang{3}, respectively \cite{patil2007improving, hetaimish2012meta}. One main reason for causing the large errors is that no real-time feedback on the orientation of the proximal tibial resection is provided \cite{iorio2013accuracy}.

In this work, we propose a robust and real-time intra-operative tibial pre-resection plane measurement algorithm for CON-TKA which could provide real-time feedback for surgeons. The algorithm uses information from a pre-operative tibia CT scans, intra-operative 2D X-ray images, and a trocar pin 3D mesh model to estimate the poses of the pins and the X-ray frames. Based on these, the resection plane can be accurately calculated.

Our work has some relations to the research works on post-operative evaluation in orthopaedic surgeries \cite{mahfouz2003robust, kim2011novel, kobayashi2009automated}, reconstruction of 3D bone surface model \cite{baka20112d, zheng20092d}, and surgical navigation \cite{otake2011intraoperative} using 2D-3D registration between pre-operative 3D CT scans and 2D X-ray images. Features (anatomical points, markers or edges), intensity or gradient information in 3D scans and X-ray images are used in the registration methods. However, these methods all have their own limitations. For example, optimising X-ray and implants separately which result in suboptimal results \cite{mahfouz2003robust, kim2011novel, kobayashi2009automated}, requiring very accurate initialisation \cite{baka20112d, zheng20092d, markelj2012review}, relying on a custom-designed hybrid fiducial for C-arm pose estimation and tracking which results in large incision \cite{otake2011intraoperative}, and requiring the time-consuming generation of digitally reconstructed radiographs (DRRs) \cite{markelj2012review}.

Some new techniques and devices have also been developed to improve the accuracy and reproducibility in maintaining proximal tibial resection, distal femoral resection, and implants alignment, including computer-assisted navigation system \cite{jones2018current}, robot-assisted surgery \cite{hampp2019robotic}, patient-specific instrumentation \cite{sassoon2015systematic}, and accelerometer-based navigation devices \cite{nam2013accelerometer}. However, they add many extra operative procedures and instruments, which increase the operation complexity, time, cost and even the risk of infections and blood clots. And no significant differences were found in mid-to-long term functional outcomes \cite{gao2019comparison}.

In summary, compared to existing methods, the key advantages of our proposed framework are (i) no external fiducials or markers are needed, (ii) pin poses and X-ray poses are jointly optimised to obtain very accurate and robust intra-operative pre-estimation of the tibial resection plane, and (iii) it can be easily integrated clinically, without interrupting the workflow of CON-TKA.

\section{Problem statement}\label{Sec_Problem_Framework}
During a standard CON-TKA surgery, the tibial cutting block is first manually aligned using an EM tibial alignment guide and secured to the patient's tibia with a pair of trocar pins, then the proximal end of the tibia will be sawed off by an oscillating saw through the deepest portion of the trochlear groove on the tibial cutting block (refer to Fig.~\ref{fig:framework}).

In the proposed framework, the aligned cutting block is removed before the execution of bone cutting and different views of X-ray images are captured for the proximal tibia with drilled pins on it using a C-arm X-ray device. The problem considered in this work is to use the pre-operative tibia mesh model, the pin mesh model, and the intra-operative X-ray observations to estimate the poses of the pins and X-ray frames w.r.t. the pre-operative tibia model intra-operatively.

Suppose $N$ views of X-ray images are captured and $k \in \{1,...,N\}$ is the index of the X-ray image, let $\mathcal{C}_{k}=\{\bar{\mathbf{p}}_{k,1},...,\bar{\mathbf{p}}_{k,\mathbb{N}_{k}}\}$ represent the edge observations of the tibia and fibula on the $k$-th X-ray image, and $\mathcal{C}_{k}^{l}=\{\bar{\mathbf{p}}^{l}_{k,1},...,\bar{\mathbf{p}}^{l}_{k,\mathbb{N}_{kl}}\}$ represent the edge observations of the $l$-th pin on the $k$-th X-ray image, $\mathbb{N}_{k}$ is the number of tibia and fibula edge pixels extracted from the $k$-th X-ray image, $l \in \left \{1,2 \right \}$ denotes the index of pin drilled into the tibia (only two pins are used in the CON-TKA procedures) and $\mathbb{N}_{kl}$ is the number of pin edge pixels extracted from the $l$-th pin in the $k$-th X-ray image.

Suppose $\mathbf{p}_{k,i}=[u_{k,i},v_{k,i}]^T$ is the ground truth coordinates of the $i$-th observed 2D edge point $\bar{\mathbf{p}}_{k,i}$, its corresponding 3D point from the pre-operative tibia model $M_{tibia}$ is $\mathbf{P}^{M}_{k,i} \in \mathbb{R}^3$, then the observation model of the tibia edge point can be written as:
\begin{equation}\label{Eq_Projection1}
	\begin{aligned}\
		\bar{\mathbf{p}}_{k,i} = \mathbf{p}_{k,i} + w_{ki}, \
		[\mathbf{p}_{k,i}^T, 1]^T\propto K (R_{k}^{{C}} \mathbf{P}^{M}_{k,i} + \mathbf{t}_{k}^{{C}})
	\end{aligned}
\end{equation}
where $w_{ki}$ is the zero-mean Gaussian noise with covariance matrix $\Sigma_{{p}_{ki}}$, $K$
is the C-arm camera intrinsic matrix, $X^C_k = \{R_{k}^{{C}}, \mathbf{t}_{k}^{{C}}\}$ represents the rotation matrix and translation vector of the C-arm camera pose at which the $k$-th X-ray image is captured (in the frame of the pre-operative tibia model).

Similarly, suppose $\bar{\mathbf{p}}^l_{k,j} =  \mathbf{p}^l_{k,j} + w_{kj}^{l} $ is the $j$-th observed 2D edge point, in which $w_{kj}^{l}$ is the zero-mean Gaussian noise with covariance matrix $\Sigma_{{p}_{kj}^{l}}$, and $\bar{\mathbf{p}}^l_{k,j}$'s corresponding 3D point from the pin model $M_{pin}$ is $\mathbf{P}^{M_l}_{k,j} \in \mathbb{R}^3$. Then, the observation model of the pin edge point is:
\begin{equation}\label{Eq_Projection2}
	\begin{aligned}\
		[(\mathbf{p}^l_{k,j})^T, 1]^T \propto K ( R_{k}^{{C}} (R_{l}^{{M}} \mathbf{P}^{M_l}_{k,j} + \mathbf{t}_{l}^{{M}}) + \mathbf{t}_{k}^{{C}})
	\end{aligned}
\end{equation}
where $X_{l}^{{M}} = \{R_{l}^{{M}}, \mathbf{t}_{l}^{{M}}\}$ is the pose of the $l$-th pin relative to the pre-operative tibia model $M_{tibia}$.
Thus, (\ref{Eq_Projection2}) first transforms $\mathbf{P}^{M_l}_{k,j}$ from the pin model frame to the pre-operative tibia model frame by using the pin pose $X_{l}^{{M}}$, and then projects onto the $k$-th camera frame by using $X^C_k$.

Mathematically, the problem considered in this paper is, given the pre-operative 3D tibia model $M_{tibia}$ and the pin model $M_{pin}$, how to accurately estimate the pin poses $\{X_{l}^{{M}}\},~ l \in \left \{1,2 \right \}$ from $N$ views of 2D intra-operative X-ray observations, where the C-arm camera poses $\{X^C_k\}, k \in \{1,...,N\}$ of the X-ray images are also not available.

\section{Resection plane estimation}
To obtain a more accurate and robust estimation, we formulate the problem as a SLAM problem where the pin poses and the C-arm camera poses are estimated simultaneously. Then, the proximal tibial pre-resection plane can be calculated using the estimated pin poses.

\subsection{SLAM formulation for pin poses estimation}
In the proposed SLAM formulation, the state to be estimated is defined as
\begin{multline}\label{State}
		\mathbf{X} \triangleq \left\{ \{X^C_k\}_{k=1}^{N}, ~ \{X_{l}^{{M}}\}_{l=1}^{2}, 
		\{\{\mathbf{P}_{k,i}^{M}\}_{i=1}^{\mathbb{N}_{k}}\}_{k=1}^{N}, \right.
		\left. \{\{\{\mathbf{P}_{k,j}^{M_l}\}_{j=1}^{\mathbb{N}_{kl}}\}_{l=1}^{2}\}_{k=1}^{N} \right \}
\end{multline}
where the variables are defined in Section \ref{Sec_Problem_Framework}.
The proposed SLAM problem can be mathematically formulated as a nonlinear optimisation problem minimising:
\begin{equation}\label{Objective}
	E = w_{rp}E_{rp} + w_{bp}E_{bp} + w_{mp}E_{mp}
\end{equation}
which consists of three energy terms: Contour Re-projection Term $E_{rp}$, Contour Back-projection Term $E_{bp}$ and Model Projection Term $E_{mp}$. $w_{rp}$, $w_{bp}$ and $w_{mp}$ are the weights of the three terms. 
Overall, the three energy terms are used together to ensure the accuracy and robustness of the estimation algorithm.

\textbf{The Contour Re-projection Term} penalises the misalignment between the contours of the outer edges of the tibia and the pins in the X-ray images and the contours re-projected by using the variables in the state in (\ref{State}):
\begin{equation}\label{Eq_Projection6}
	\begin{aligned}
		&E_{rp} = \sum_{k=1}^{N}  \left [
		\sum_{i=1}^{\mathbb{N}_{k}}
		\left \| \bar{\mathbf{p}}_{k,i}  - \mathbf{p}_{k,i} \right \|^{2}_{\Sigma_{{p}_{ki}}^{-1}}  + 
		\sum_{l=1}^{2}  \sum_{j=1}^{\mathbb{N}_{kl}}  \left\| \bar{\mathbf{p}}_{k,j}^l -  \mathbf{p}_{k,j}^l \right \|^{2}_{\Sigma_{{p}_{kj}^l}^{-1}} \right ]
	\end{aligned}
\end{equation}
where $\Sigma_{{p}_{ki}}$ and $\Sigma_{{p}_{kj}^l}$ are the covariance matrices of the 2D contour feature observations $\bar{\mathbf{p}}_{k,i}$ and $\bar{\mathbf{p}}^l_{k,j}$, respectively. $\mathbf{p}_{k,i}$ and $\mathbf{p}_{k,j}^l$ are given in (\ref{Eq_Projection1}) and (\ref{Eq_Projection2}).

\textbf{The Contour Back-projection Term} minimises the sum of squared coordinate distances (along the $x$, $y$ and $z$ axes, respectively) between the pre-operative tibia model $M_{tibia}$ and the 3D back-projected contour points $\{\mathbf{P}_{k,i}^{M}\}$ in the state $\mathbf{X}$, and between the pin model $M_{pin}$ and the 3D back-projected contour points $\{\mathbf{P}_{k,j}^{M_l}\}$ in the state $\mathbf{X}$:
\begin{multline}\label{Eq_Projection9}
	E_{bp} = \sum_{k=1}^{N}  \left [
	\sum_{i=1}^{\mathbb{N}_{k}} 
	\left \| \mathbf{d}(\mathbf{P}_{k,i}^{M},M_{tibia}) \right \|^{2}_{\Sigma_{{P}_{ki}^{M}}^{-1}} + \right.
	\left. \sum_{l=1}^{2}  \sum_{j=1}^{\mathbb{N}_{kl}} \left\| \mathbf{d}(\mathbf{P}_{k,j}^{M_l},M_{pin}) \right \|^{2}_{\Sigma_{{P}_{kj}^{M_l}}^{-1}} \right ]
\end{multline}
where $\mathbf{d}(\mathbf{P}_{k,i}^{M},M_{tibia})$ represents the shortest coordinate distance from 3D point $\mathbf{P}_{k,i}^{M}$ to $M_{tibia}$, and $\mathbf{d}(\mathbf{P}_{k,j}^{M_l},M_{pin})$ represents the shortest coordinate distance from $\mathbf{P}_{k,j}^{M_l}$ to $M_{pin}$, $\Sigma_{{P}_{ki}^{M}}$ and $\Sigma_{{P}_{kj}^{M_l}}$ are the corresponding covariance matrices.

\textbf{The Model Projection Term} penalises the misalignment between the observed contours in the intra-operative X-ray images and contours extracted from the projection of the pre-operative tibia model $M_{tibia}$ and pin model $M_{pin}$:
\begin{equation}\label{model_projection}
	\begin{aligned}
		E_{mp}  =  \sum_{k=1}^{N}   \left [
		\sum_{m=1}^{\mathbb{N}^M_{k}}
		\left \| d(\tilde{\mathbf{p}}_{k,m}, \mathcal{C}_{k}) \right \|^{2}_{\Sigma_{{p}_{km}}^{-1}}  +  
		\sum_{l=1}^{2}  \sum_{n=1}^{\mathbb{N}^M_{kl}}  \left\| d( \tilde{\mathbf{p}}_{k,n}^l, \mathcal{C}^{l}_{k}) \right \|^{2}_{\Sigma_{{p}_{kn}^l}^{-1}}  \right ]
	\end{aligned}
\end{equation}
where $\{\tilde{\mathbf{p}}_{k,m}\}$, $m \in\{1,...,\mathbb{N}^M_{k}\}$  are the 2D contour points representing the edge of the pre-operative tibia model projection, projected from $M_{tibia}$ to the $k$-th X-ray image using $X^C_k$. $d(\tilde{\mathbf{p}}_{k,m}, \mathcal{C}_{k})$ means the shortest Euclidean distance between $\tilde{\mathbf{p}}_{k,m}$ and the contour $\mathcal{C}_{k}$ of the $k$-th X-ray image.
Similarly, $\{\tilde{\mathbf{p}}_{k,n}^l\}$, $n \in\{1,...,\mathbb{N}^M_{kl}\}$ are the 2D contour points representing the edge of the $l$-th pin projection, projected from $M_{pin}$ to the $k$-th X-ray image using $X^C_k$ and $X^M_l$, $d( \tilde{\mathbf{p}}_{k,n}^l, \mathcal{C}^{l}_{k})$ is the shortest Euclidean distance between $\tilde{\mathbf{p}}_{k,n}^l$ and the contour $\mathcal{C}^{l}_{k}$.

\subsection{Resection plane estimation and the approach overview}\label{Sec_Framework}
\begin{figure*}[htbp!]
	\centering
	\includegraphics[width=1.0\textwidth]{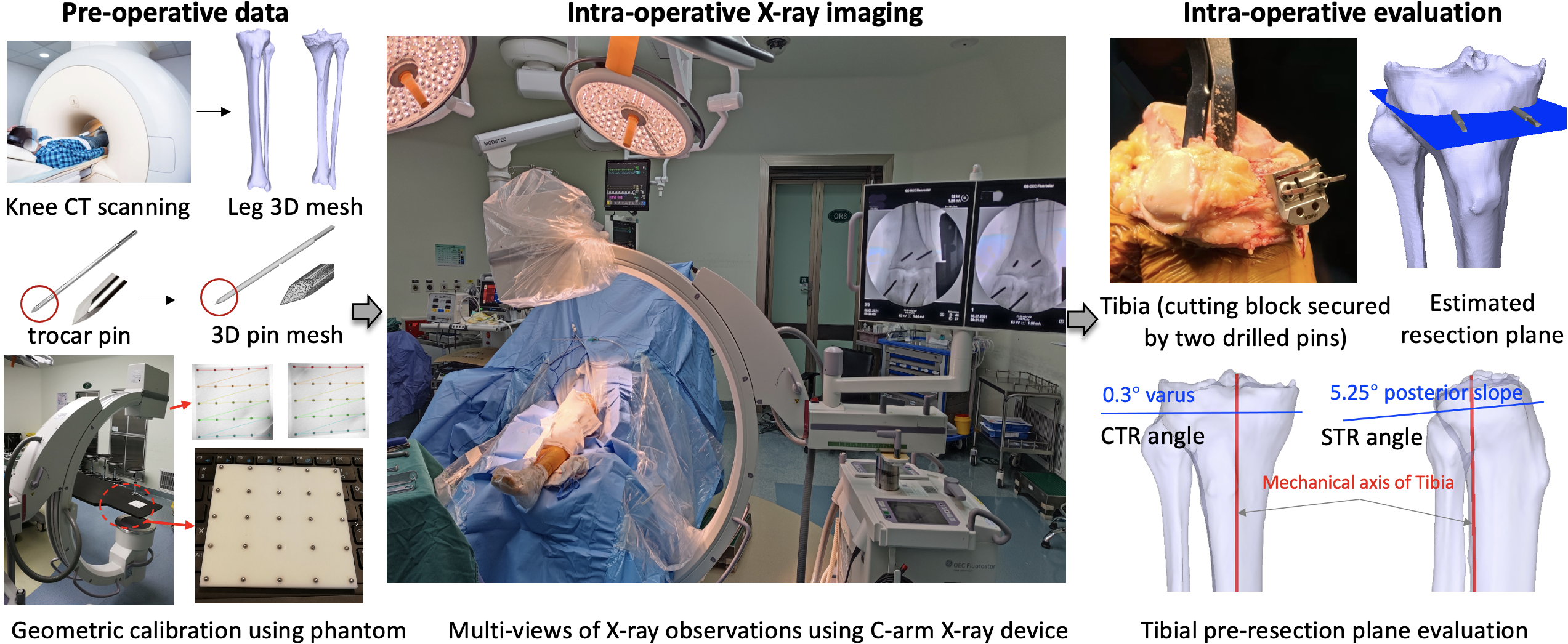}
	\caption{The framework of proposed proximal tibial resection plane estimation.}
	\label{fig:framework}
\end{figure*}
The iterative Gauss-Newton (GN) algorithm is used to minimise the objective function (\ref{Objective}).
After the optimisation, the two pins are transformed into the frame of the pre-operative tibia model using the estimated poses, and the corresponding tibial resection plane (parallel to the plane fitted by the two paralleled trocar pins) is obtained by fitting all the vertices of the two transformed pin models. The coronal tibial resection (CTR) angle and the sagittal tibial resection (STR) angle of the estimated tibial resection plane are calculated w.r.t. the tibial mechanical axis and compared to the desired bone resection requirement before the surgeon cuts the proximal end of the tibia.
Fig.~\ref{fig:framework} shows the proposed framework which consists of pre-operative data preparation, intra-operative X-ray imaging, and intra-operative tibial resection plane estimation and evaluation.

\section{Experiments}
The proposed SLAM-TKA framework for real-time estimating the proximal tibial pre-resection plane is validated by simulations and in-vivo experiments.
The pre-operative 3D tibia model is segmented from CT scans using the Mimics software, the 3D trocar pin mesh model is created using the Solidworks software. 
The minimum number of X-ray images are used both in simulations and in-vivo experiments, then $N=2$.
The C-arm X-ray device was calibrated using a calibration phantom, and the calibration dataset and code are made available in GitHub (\url{https://github.com/zsustc/Calibration.git}).

The proposed framework is compared to the feature-based 2D-3D registration ``projection'' strategy used in \cite{mahfouz2003robust, kim2011novel, kobayashi2009automated} and the ``back-projection'' strategy used in \cite{zheng20092d, baka20112d}.
Four methods are compared, namely ``projection (split)'', ``projection (joint) '', ``back-projection (split)' and ``back-projection (joint)''. ``split'' means the poses of X-ray frames and pins are estimated separately, ``joint'' means all X-ray frames and pins are optimised together. We have implemented the compared methods ourselves since the source codes of the compared works are not publicly available.
Given that there is practically no pixel-wise correspondences between the generated DRRs of the pin mesh model and X-ray images \cite{markelj2012review}, currently the DRR-based 2D-3D registration method used in \cite{otake2011intraoperative} is not compared.

\subsection{Simulation and robustness assessment}
\label{sec:simulation_exp}
\begin{figure*}[hpbt!]
	\centering
	\begin{subfigure}[t]{.495\textwidth}
		\centering
		\includegraphics[width=1.0\textwidth]{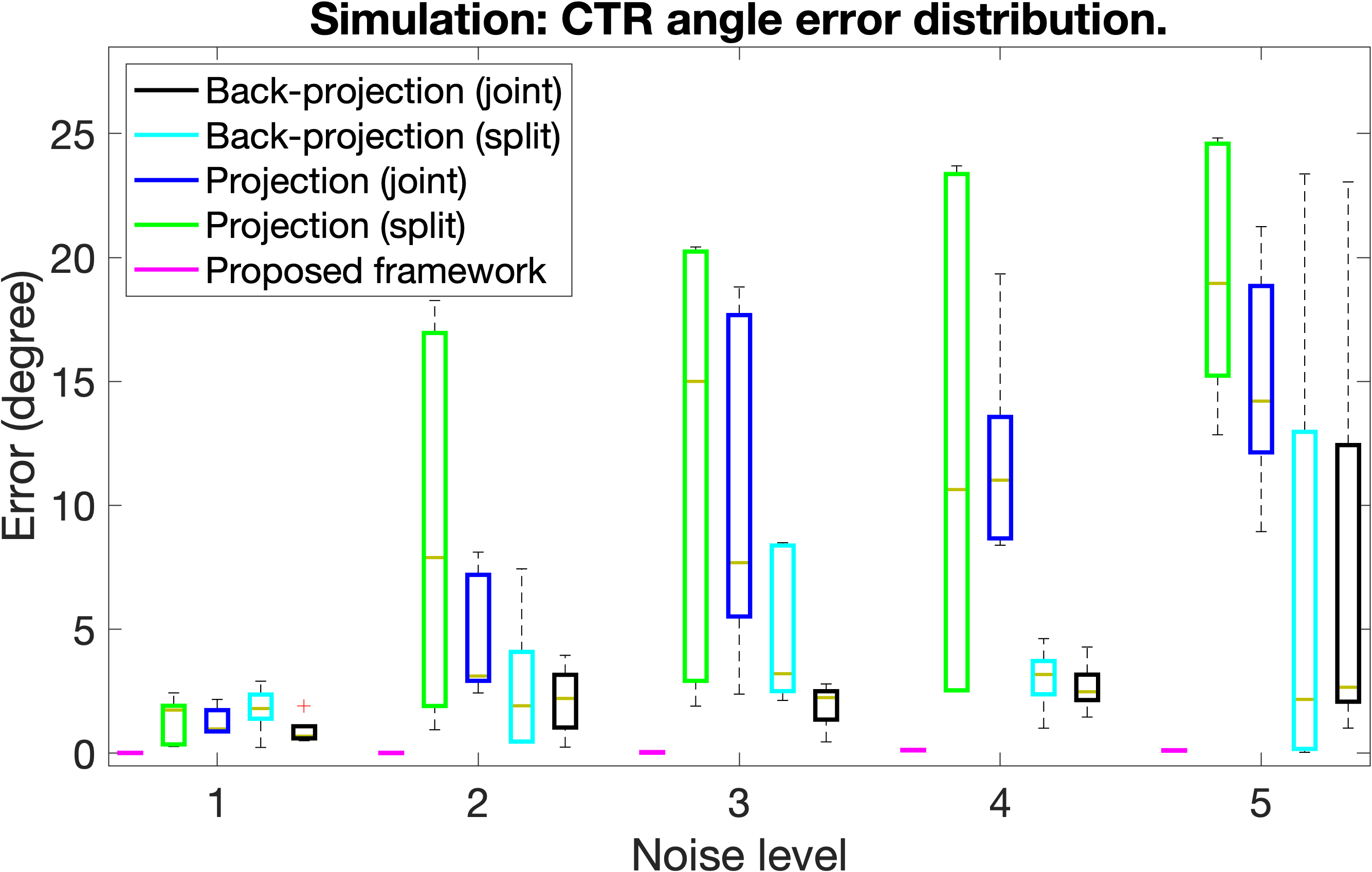}
	\end{subfigure}
	\begin{subfigure}[t]{.495\textwidth}
		\centering
		\includegraphics[width=1.\textwidth]{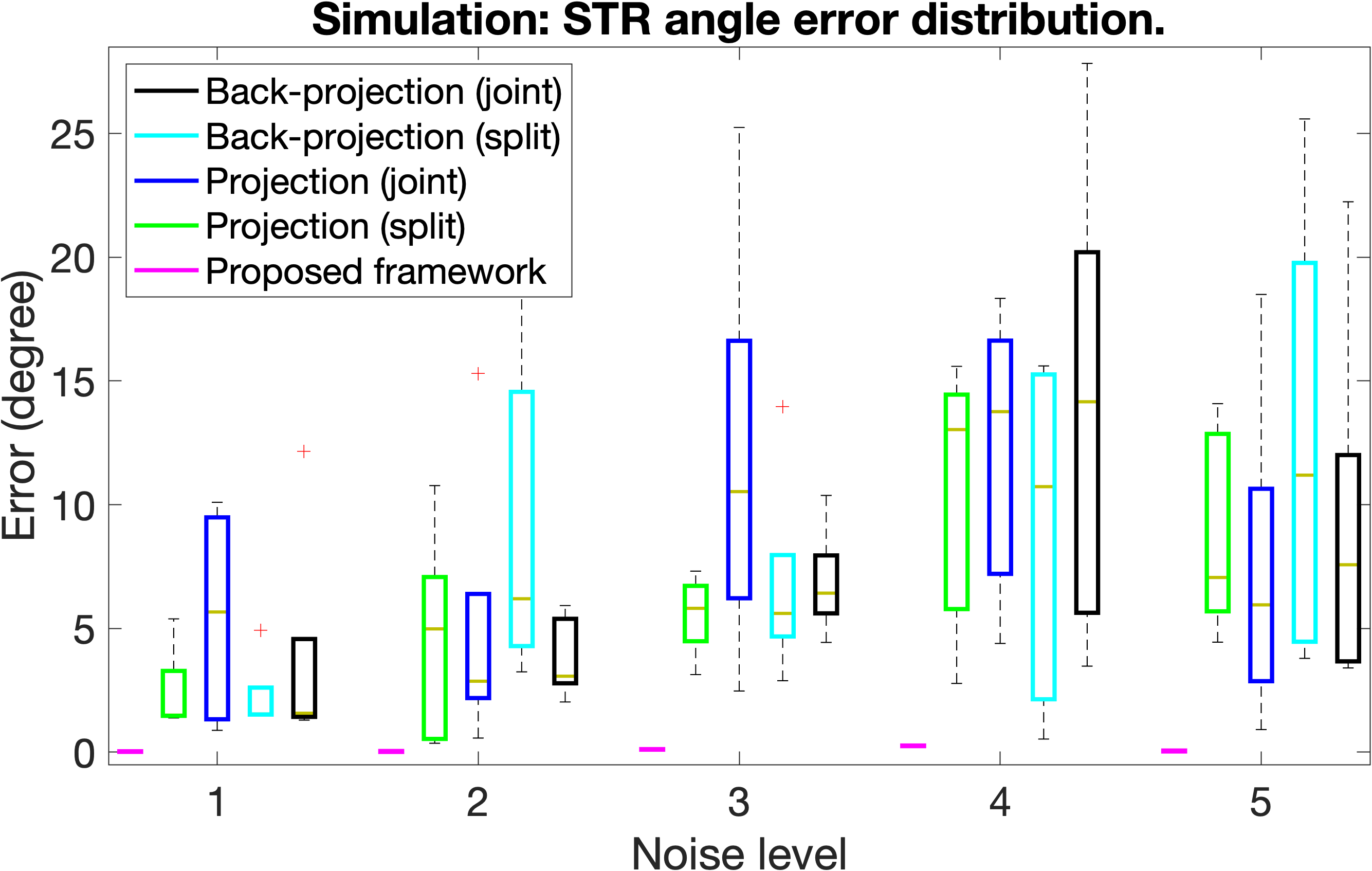}
	\end{subfigure} 
	\caption{Resection plane error for the simulation with five increasing noise levels.}
	\label{fig:simulation_error}
\end{figure*}
To simulate the in-vivo scenario, zero-mean Gaussian noise with standard deviation (SD) of $2$ pixels is added to the edge feature observations on the X-ray images. Five different levels of noise are added to the ground truth and used in the initialisation of state $\bm X$, zero-mean Gaussian noise with SD of $\left\{0.1, 0.2, 0.3, 0.4, 0.5\right\}$ rad is added to the rotation angles, and noise with SD of $\left\{20, 40, 60, 80, 100\right\}$ mm is added to the translations and all the 3D edge points.

Ten independent runs are executed for each noise level to test the robustness and accuracy of the proposed algorithm and the compared methods.
The absolute errors of CTR and STR angles (compared to the ground truth of the simulated resection plane) of the ten runs for the estimation of tibial resection plane are shown in Fig.~\ref{fig:simulation_error}. 
It shows that the proposed framework has the highest accuracy and robustness in terms of CTR and STR angles. 

\subsection{In-vivo experiments}
\label{sec:invivo_exp}

We use data from five CON-TKAs for evaluating the proposed algorithm.
The doctor uses a stylus together with a touch screen to draw and extract the contour edges of the proximal tibia and pins from the intra-operative X-ray images, and only clear tibia and fibula contours are used as observations.
The state variables $\{X^C_k\}$ are initialised by the C-arm rotation angles and translation measured using the device joint encoders, and $\{X^M_l\}$ are initialised by the pose of the ideal resection plane in the pre-operative tibia model according to the optimal tibial resection requirement. The state variables ${\textbf{P}_{k,i}^{M}}$ are initialised by back-projecting their corresponding 2D contour observation points $\bar{\mathbf{p}}_{k,i}$ into the pre-operative tibia model frame using $\{X^C_k\}$ and the initial guesses for the point depths. And the points depth can be initialised by using the distance between the patella of the patient and the X-ray tube from the C-arm X-ray device. Similar method is used to initialise the state variables ${\textbf{P}_{k,j}^{M_l}}$.

\begin{figure*}[hpbt!]
	\centering
	\begin{subfigure}[t]{.188\textwidth}
		\centering
		\includegraphics[width=1.0\textwidth]{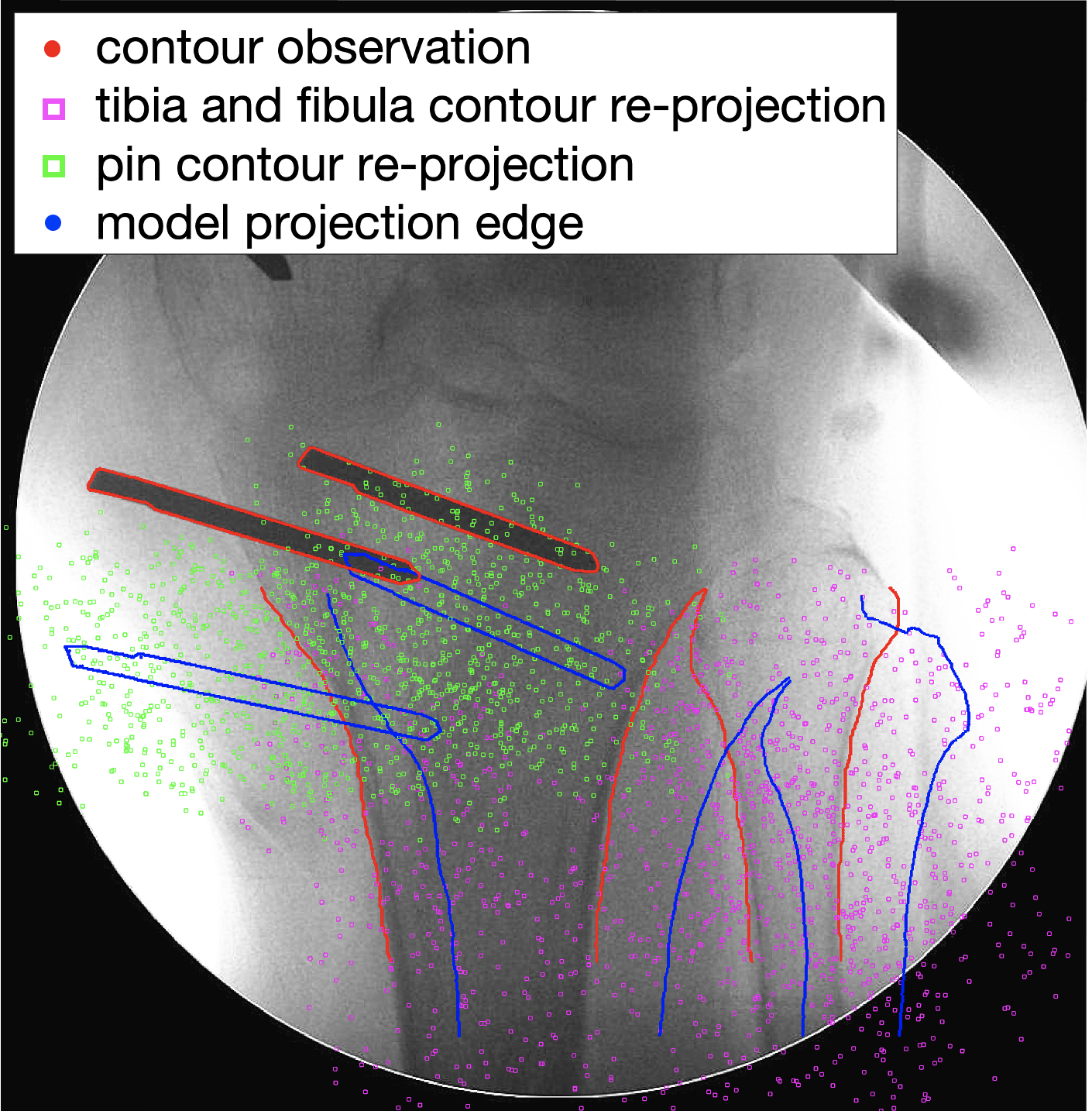} 
		\caption{Ini-2D-1st}
	\end{subfigure}
	\begin{subfigure}[t]{.194\textwidth}
		\centering
		\includegraphics[width=1.0\textwidth]{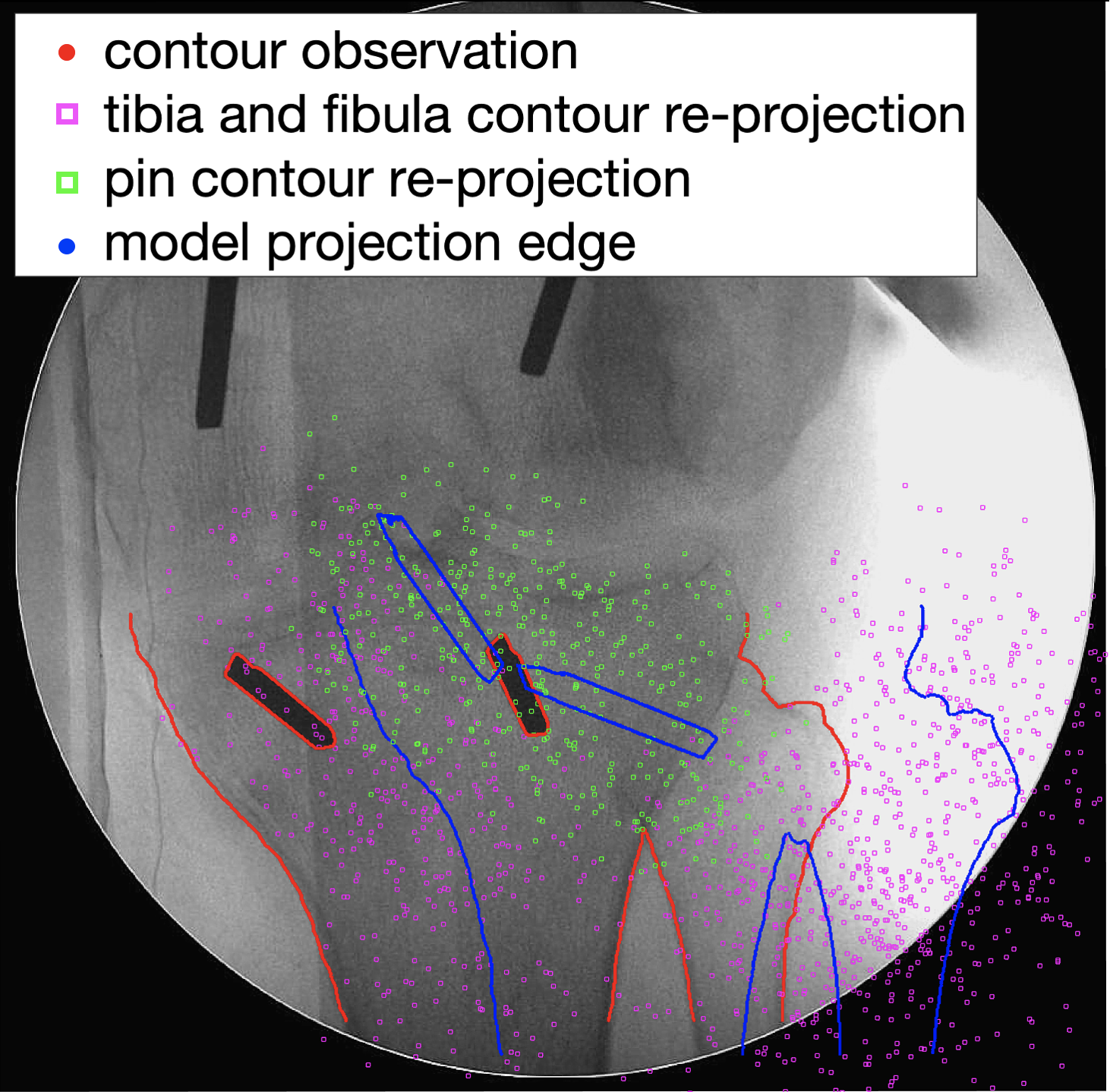} 
		\caption{Ini-2D-2nd}
	\end{subfigure} 
	\begin{subfigure}[t]{.169\textwidth}
		\centering
		\includegraphics[width=1.0\textwidth]{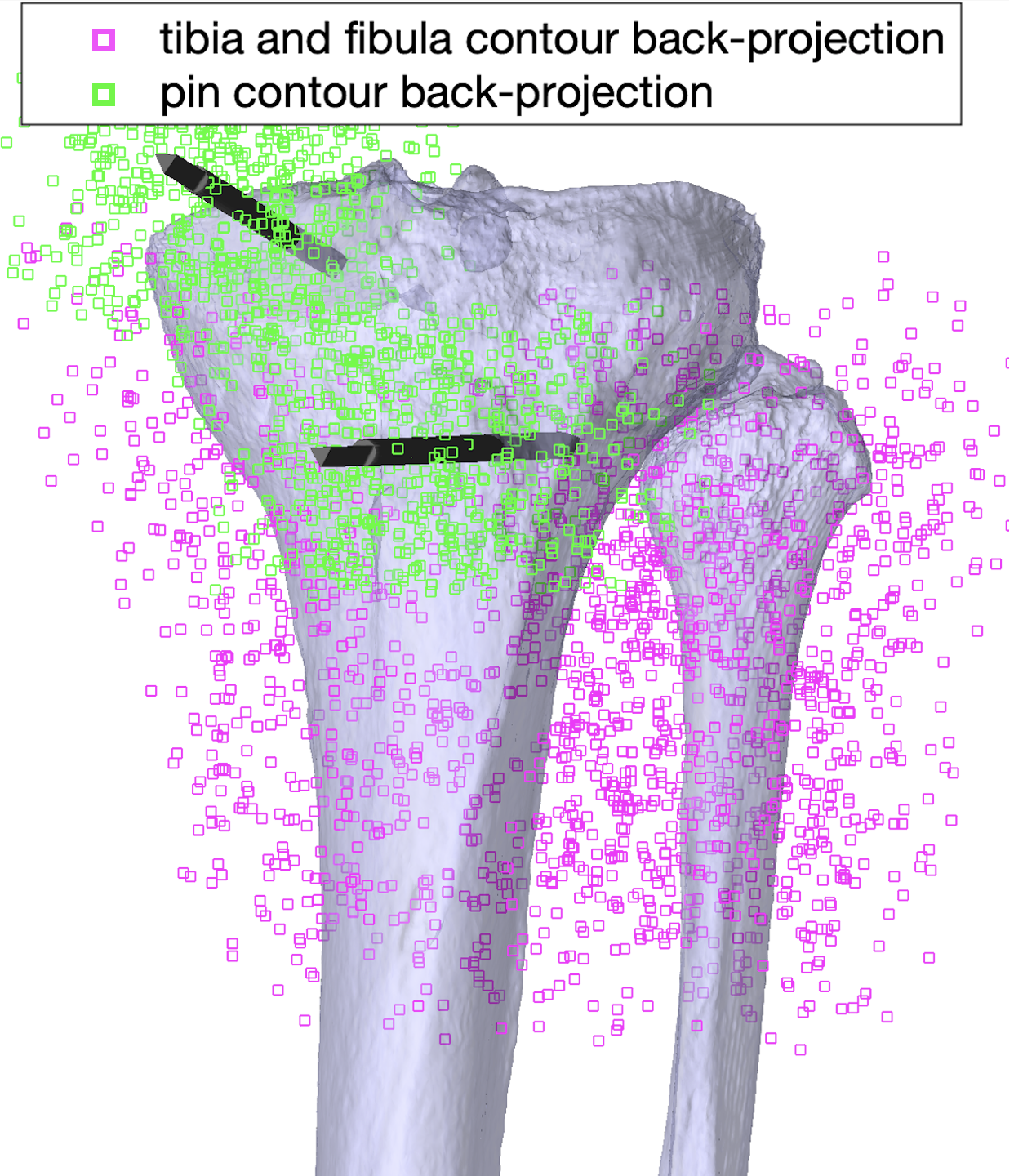} 
		\caption{Ini-3D-1st}
	\end{subfigure}
	\begin{subfigure}[t]{.168\textwidth}
		\centering
		\includegraphics[width=1.0\textwidth]{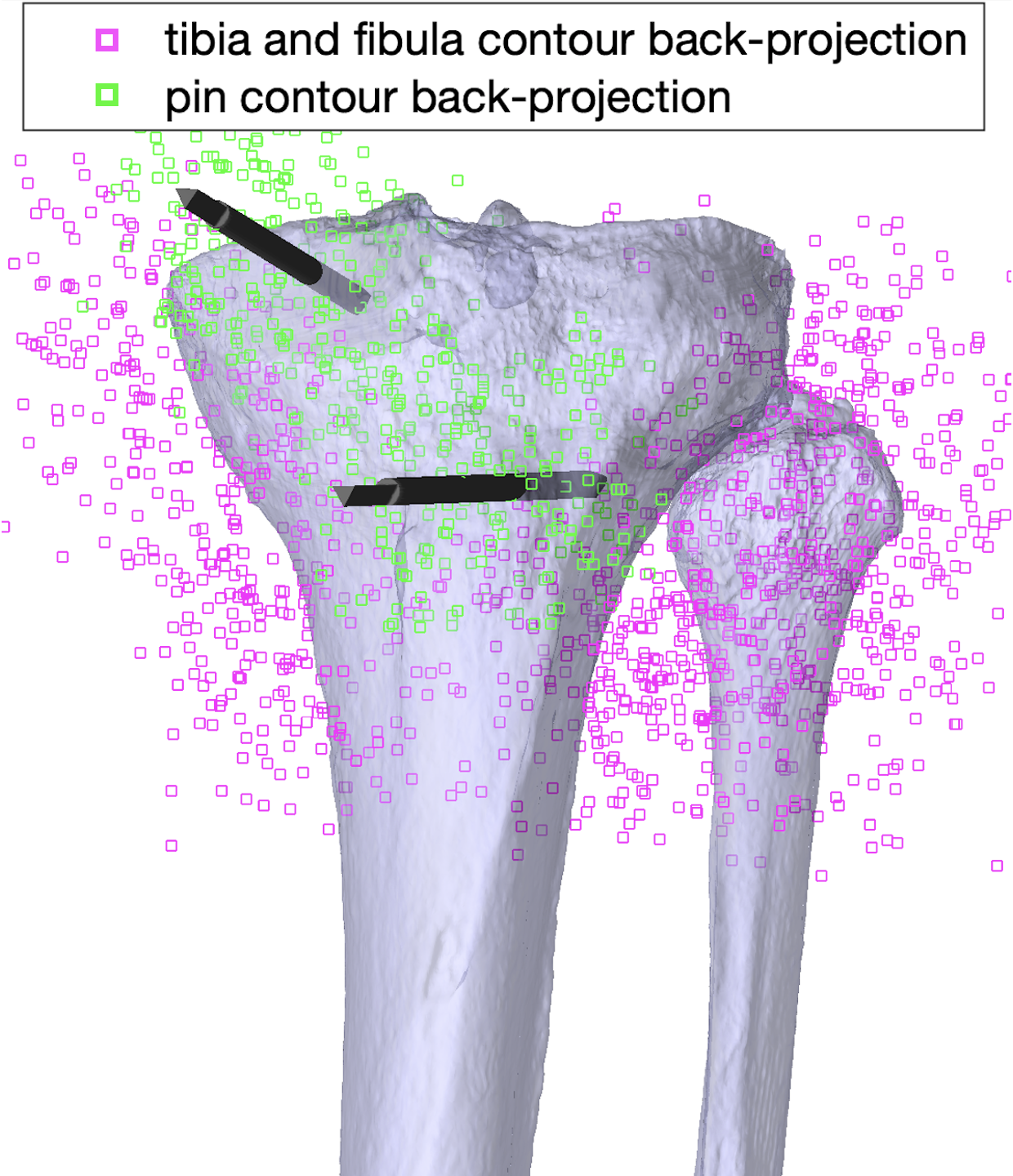} 
		\caption{Ini-3D-2nd}
	\end{subfigure}
	\begin{subfigure}[t]{.165\textwidth}
		\centering
		\includegraphics[width=1.0\textwidth]{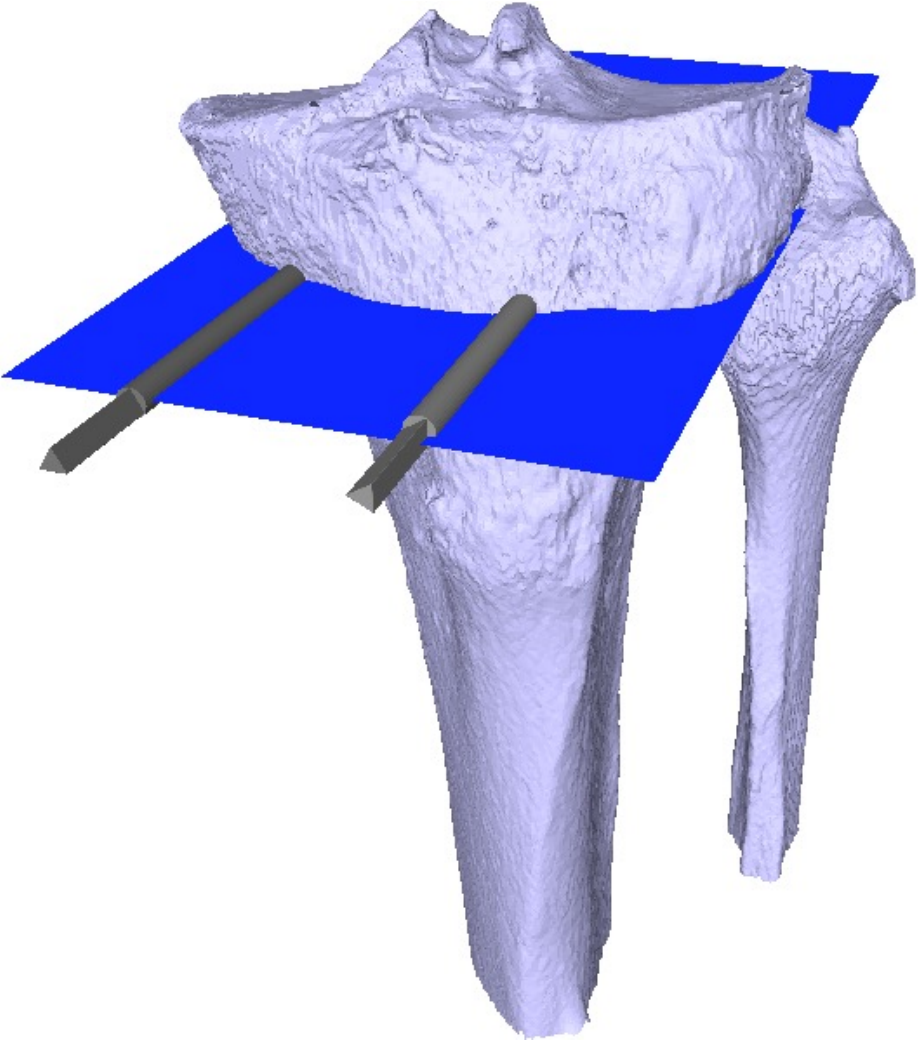} 
		\caption{1st-view}
	\end{subfigure}\\
	\begin{subfigure}[t]{.188\textwidth}
		\centering
		\includegraphics[width=1.0\textwidth]{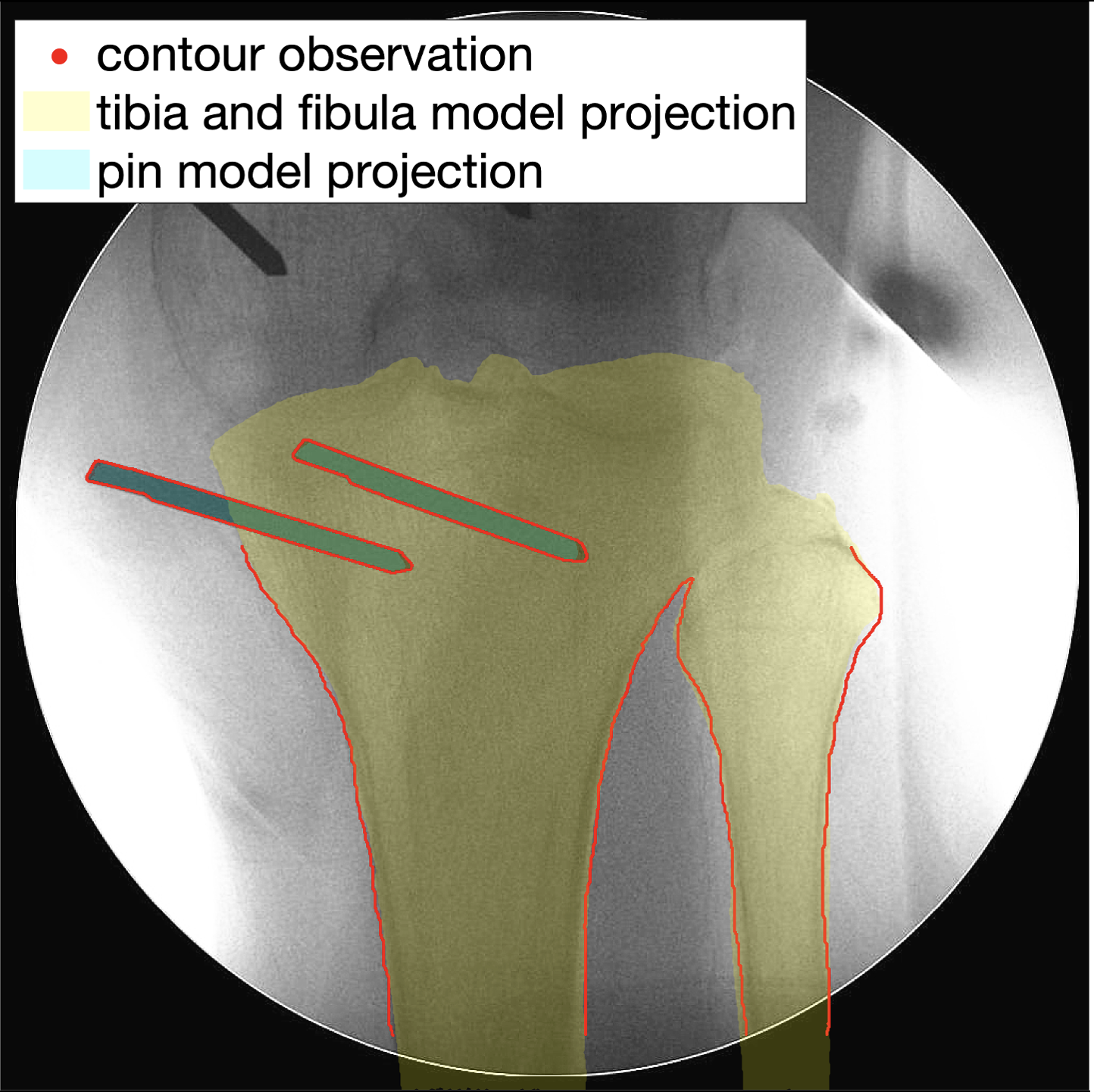} 
		\caption{Opt-2D-1st}
	\end{subfigure}
	\begin{subfigure}[t]{.188\textwidth}
		\centering
		\includegraphics[width=1.0\textwidth]{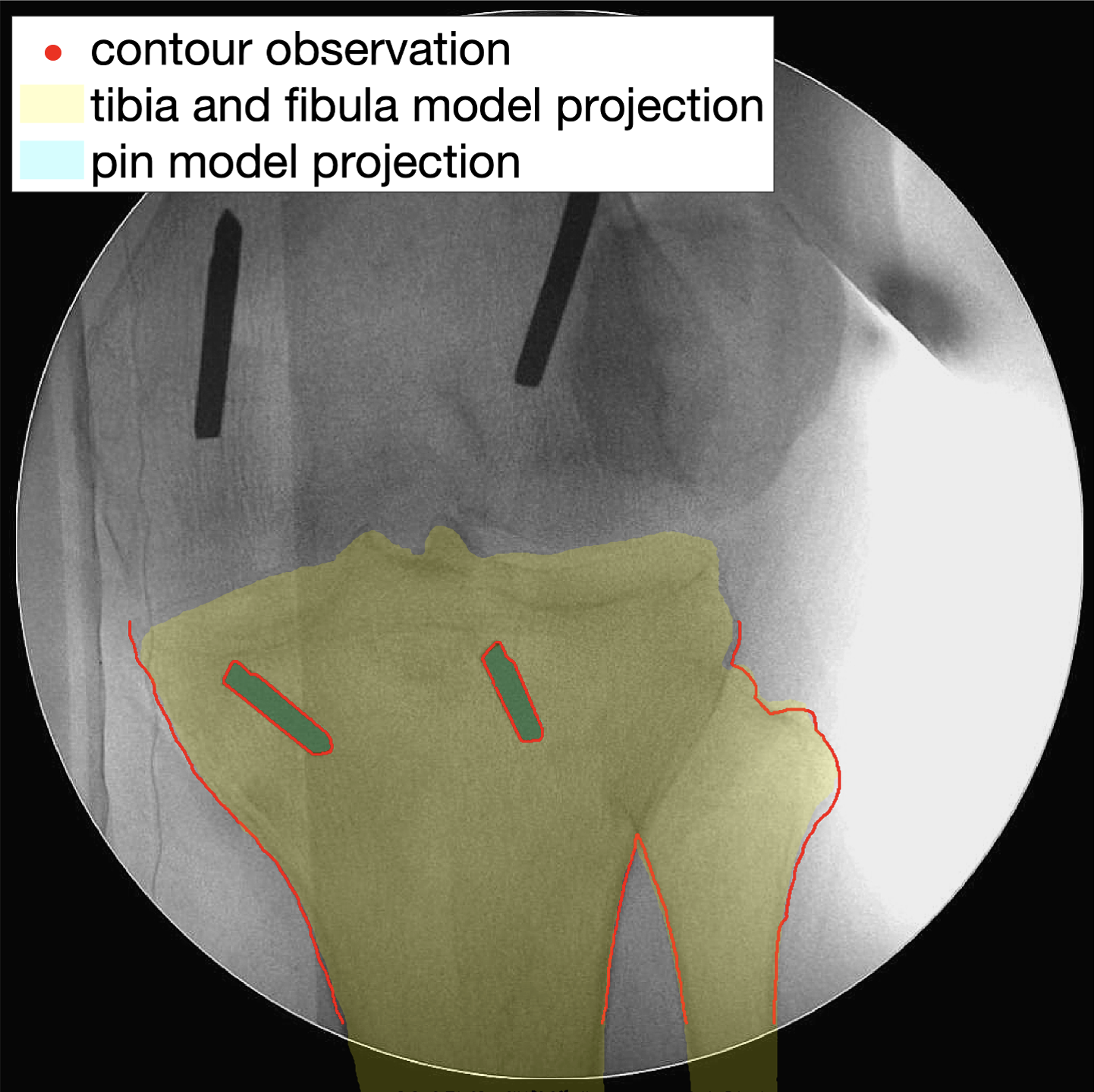}
		\caption{Opt-2D-2nd}
	\end{subfigure}
	\begin{subfigure}[t]{.178\textwidth}
		\centering
		\includegraphics[width=1.0\textwidth]{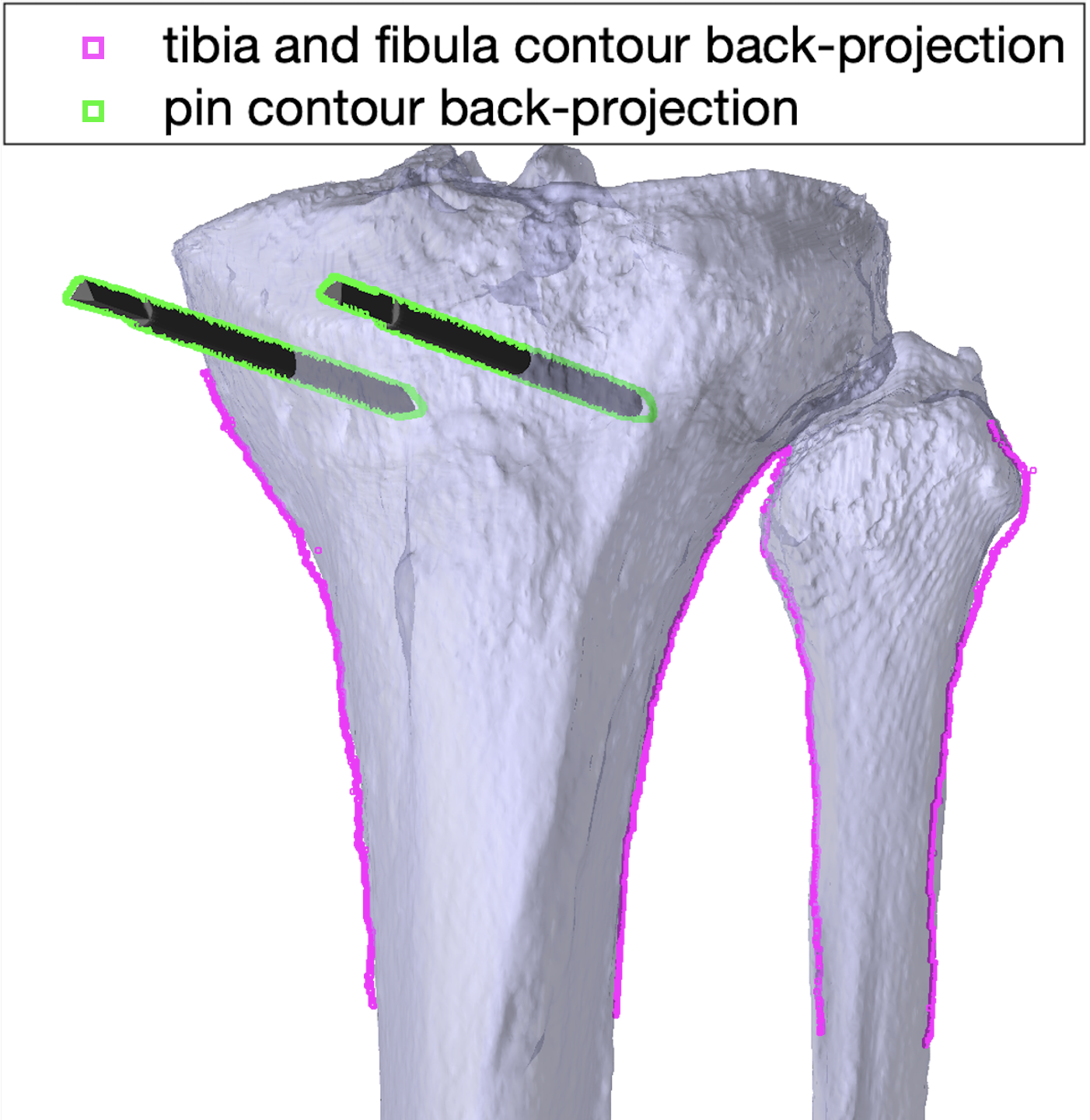} 
		\caption{Opt-3D-1st}
	\end{subfigure}
	\begin{subfigure}[t]{.175\textwidth}
		\centering
		\includegraphics[width=1.0\textwidth]{figures/figure03_opt_3d_fv} 
		\caption{Opt-3D-2nd}
	\end{subfigure}
	\begin{subfigure}[t]{.168\textwidth}
		\centering
		\includegraphics[width=1.0\textwidth]{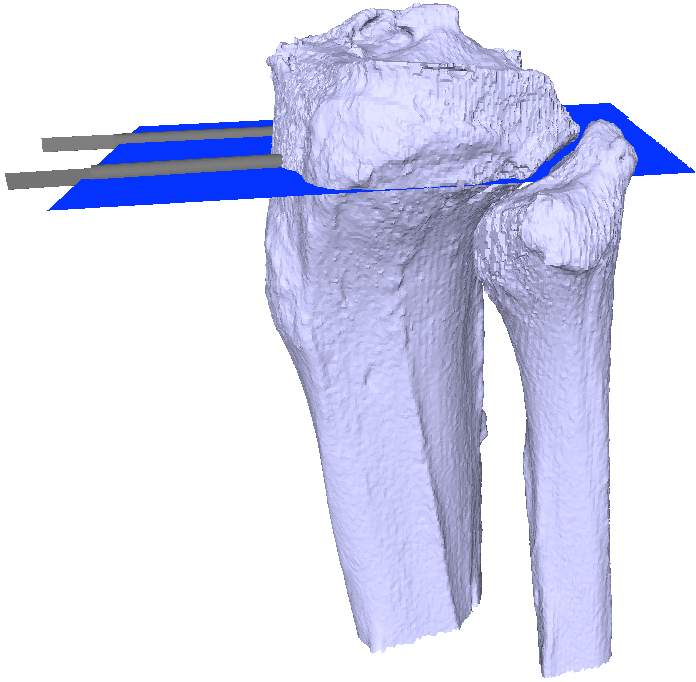} 
		\caption{2nd-view}
	\end{subfigure}
	\caption{The initialisation and optimisation on the first in-vivo dataset. `Ini'' and  ``Opt'' are the abbreviations for initialisation and optimisation, respectively. }
	\label{fig:hejianwen}
\end{figure*}

Fig.~\ref{fig:hejianwen} shows the initialisation and optimisation results of the proposed algorithm on the first in-vivo dataset. Fig.~\ref{fig:hejianwen}(a) - (d) show the tibia and pin contour re-projections onto the 2D X-ray images and contour back-projections into the pre-operative tibia model frame and the pin model frame using the initial values. It is obvious that the projected and back-projected contour features have large errors. After optimisation, referring to Fig.~\ref{fig:hejianwen}(f) - (i), the projection from the pre-operative tibia model and the pin model can fit well with the extracted observation, and the contour back-projection features are very close to the outer surface of the pre-operative tibia model and the pin model. Fig.~\ref{fig:hejianwen}(e) and Fig.~\ref{fig:hejianwen}(j) show two different views of the estimated tibial resection plane.

\begin{figure*}[htbp!]
	\centering
	\begin{subfigure}[t]{.182\textwidth}
		\centering
		\includegraphics[width=1.0\textwidth]{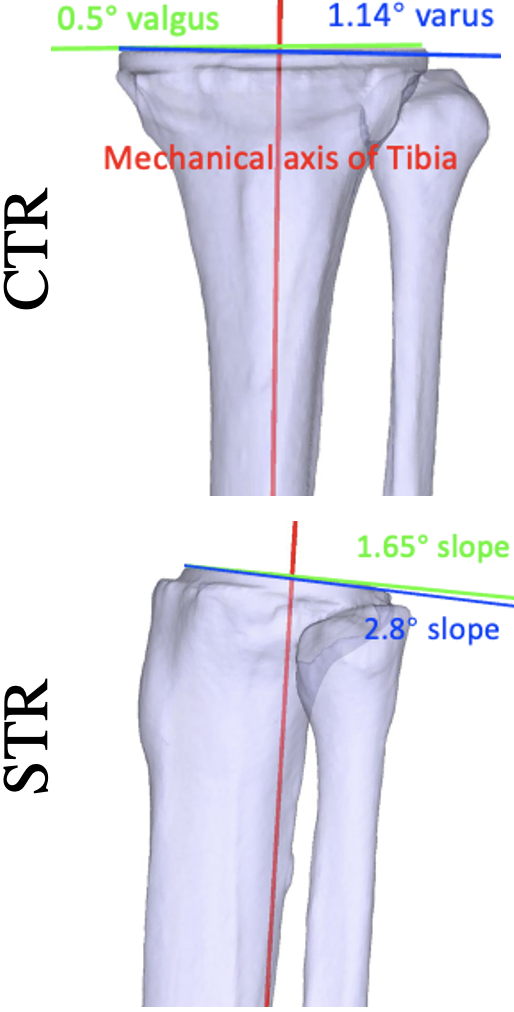}
		\caption{1st dataset}
	\end{subfigure}
	\begin{subfigure}[t]{.18\textwidth}
		\centering
		\includegraphics[width=1.0\textwidth]{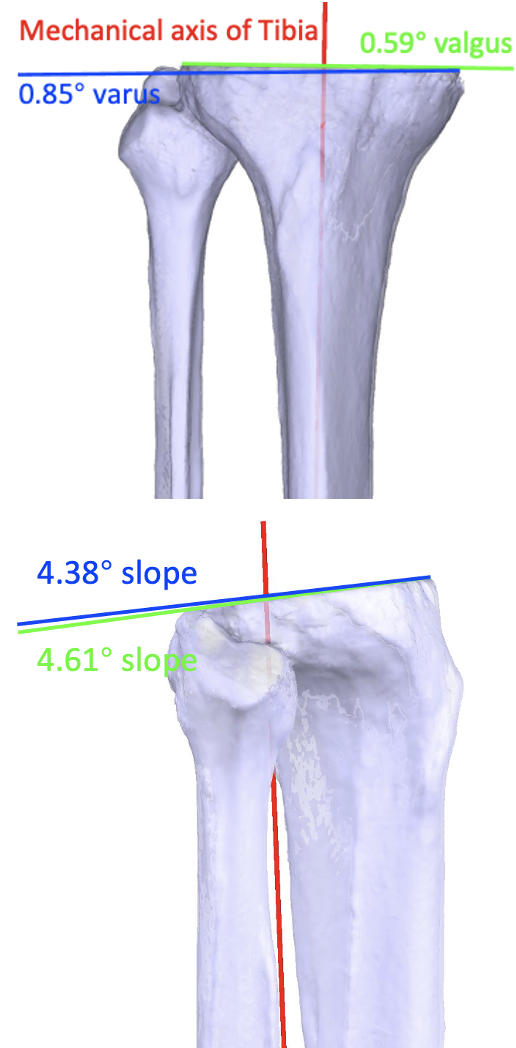}
		\caption{2nd dataset}
	\end{subfigure}
	\begin{subfigure}[t]{.175\textwidth}
		\centering
		\includegraphics[width=1.0\textwidth]{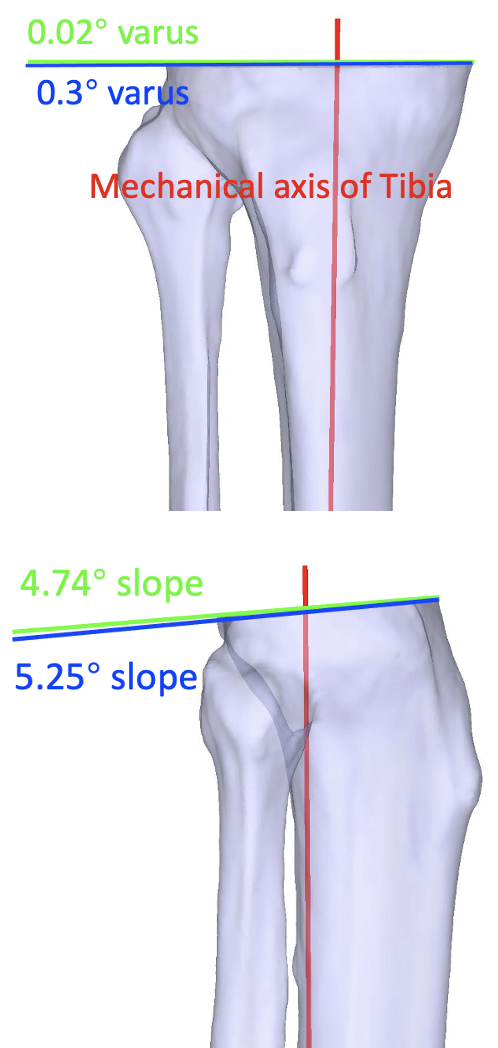}
		\caption{3rd dataset}
	\end{subfigure}
	\begin{subfigure}[t]{.173\textwidth}
		\centering
		\includegraphics[width=1.0\textwidth]{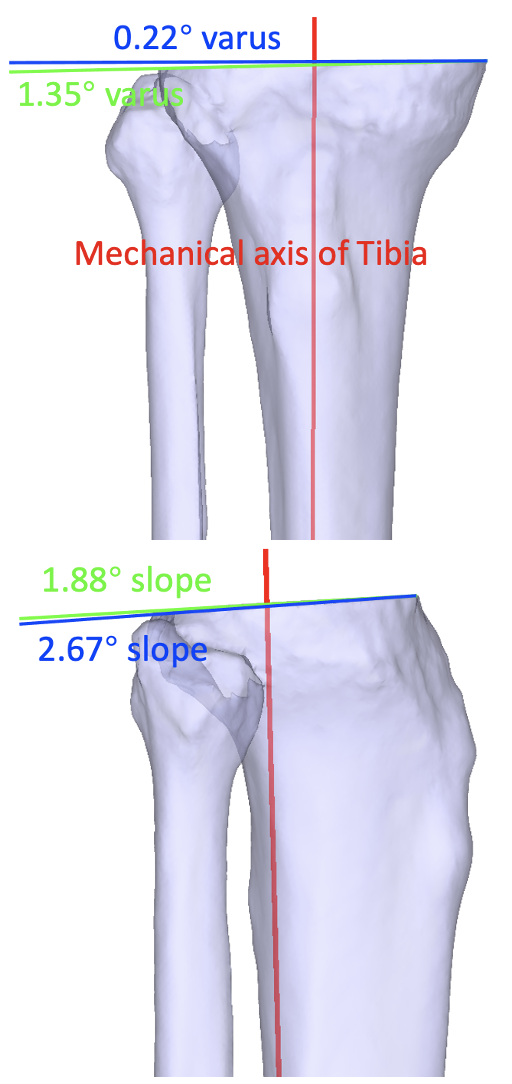}
		\caption{4th dataset}
	\end{subfigure}
	\begin{subfigure}[t]{.172\textwidth}
		\centering
		\includegraphics[width=1.0\textwidth]{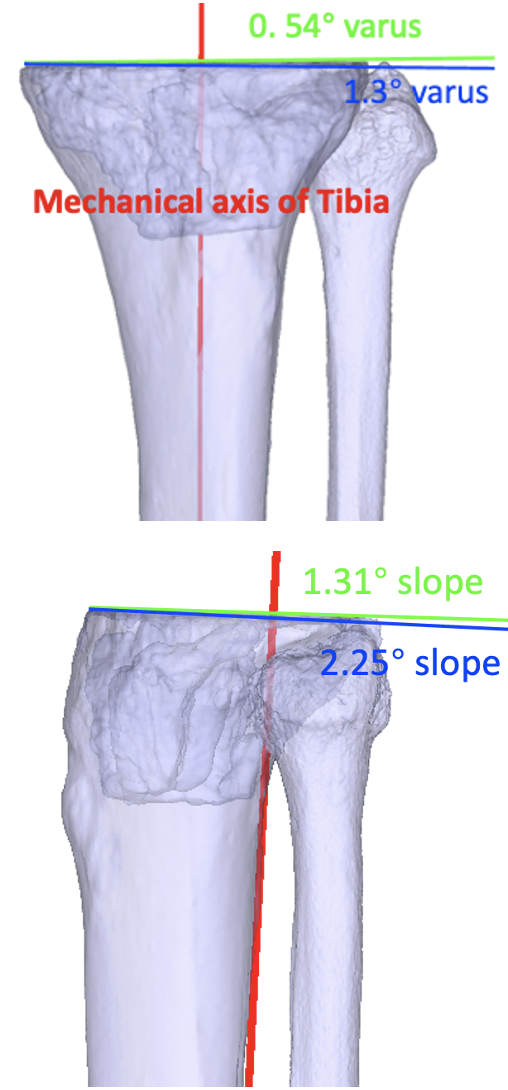}
		\caption{5th dataset}
	\end{subfigure}
	\caption{Results from the proposed framework on the five in-vivo datasets.}
	\label{fig:invivo_exp}
\end{figure*}

Fig.~\ref{fig:invivo_exp} shows the comparison between the estimated tibial resection plane angles (blue) using the proposed method and the ground truth (green) obtained from post-operative tibial CT scans for the five in-vivo datasets. The proposed method can achieve high accurate estimation of the tibial resection plane angles intra-operatively, and this can guarantee the accuracy of tibial resection by comparing the estimation to the clinical bone resection requirement (CTR and STR angles within ±3° w.r.t. the mechanical axis) before the surgeon cut the proximal end of the tibia.

\begin{table}[htbp!]
	\centering
	\caption{Resection plane estimation error in five in-vivo testings.}
	\label{tab:my-table}
	\begin{tabular}{|c|cc|cc|cc|cc|cc|}
		\hline
		Invivo datasets                    & \multicolumn{2}{c|}{1}                             & \multicolumn{2}{c|}{2}                              & \multicolumn{2}{c|}{3}                             & \multicolumn{2}{c|}{4}                              & \multicolumn{2}{c|}{5}                             \\ \hline
		Errors (\degree) & \multicolumn{1}{c|}{CTR}           & STR           & \multicolumn{1}{c|}{CTR}           & STR            & \multicolumn{1}{c|}{CTR}           & STR           & \multicolumn{1}{c|}{CTR}            & STR           & \multicolumn{1}{c|}{CTR}           & STR           \\ \hline
		Proposed framework             & \multicolumn{1}{c|}{\textbf{1.64}} & \textbf{1.15} & \multicolumn{1}{c|}{\textbf{1.44}} & \textbf{-0.23} & \multicolumn{1}{c|}{\textbf{0.28}} & \textbf{0.51} & \multicolumn{1}{c|}{\textbf{-1.13}} & \textbf{0.79} & \multicolumn{1}{c|}{\textbf{0.76}} & \textbf{0.94} \\ \hline
		Projection (split)             & \multicolumn{1}{c|}{3.72}          & 6.15          & \multicolumn{1}{c|}{3.43}          & -1.6           & \multicolumn{1}{c|}{4.13}          & 7.43          & \multicolumn{1}{c|}{-4.21}          & -4.9          & \multicolumn{1}{c|}{5.8}           & 4.0             \\ \hline
		Projection (joint)             & \multicolumn{1}{c|}{2.75}          & 5.32          & \multicolumn{1}{c|}{2.61}          & -1.78          & \multicolumn{1}{c|}{2.24}          & 8.46          & \multicolumn{1}{c|}{-3.81}          & -7.11         & \multicolumn{1}{c|}{6.23}          & 3.51          \\ \hline
		Back-projection (split)        & \multicolumn{1}{c|}{5.75}          & 4.97          & \multicolumn{1}{c|}{2.0}           & -2.89          & \multicolumn{1}{c|}{5.58}          & 8.04          & \multicolumn{1}{c|}{-5.29}          & -5.45         & \multicolumn{1}{c|}{6.75}          & 4.67          \\ \hline
		Back-projection (joint)        & \multicolumn{1}{c|}{3.51}          & 5.73          & \multicolumn{1}{c|}{3.16}          & -1.31          & \multicolumn{1}{c|}{3.97}          & 7.80          & \multicolumn{1}{c|}{-4.84}          & -4.9          & \multicolumn{1}{c|}{5.14}          & 4.18          \\ \hline
	\end{tabular}
\end{table}

Table \ref{tab:my-table} summarises the accuracy of the proposed algorithm comparing to other four methods using the five groups of in-vivo datasets, and overall the proposed framework achieves the highest accuracy. The proposed algorithm converged in about 15-20 iterations with around 1.5-2.5 seconds per iteration. 
Thus, the computation time is short enough such that the tibial resection plane estimation can be completed without much influence on the CON-TKA procedure. 
In comparison, the compared methods use numerical optimisation tools such as simulated annealing (SA) algorithm \cite{kirkpatrick1983optimization} or the Nelder-Mead Downhill Simplex algorithm \cite{nelder1965simplex} and take hours to converge.

\section{Conclusion}
This paper presents SLAM-TKA, a real-time algorithm for reliably and intra-operatively estimating the tibial resection plane for CON-TKAs. It solves a SLAM problem using a patient-specific pre-operative tibia CT scans, a trocar pin mesh model and two intra-operative X-ray images. Simulation experiments demonstrate the robustness and accuracy of the proposed algorithm. In-vivo experiments demonstrate the feasibility and practicality of TKA-SLAM.
In conclusion, the algorithm proposed in this paper can be deployed to improve tibial resection intra-operatively without the changing on the CON-TKA procedures.
In the future, we aim to use a general tibia model together with different views of intra-operative X-ray images to recover the patient-specific tibia model, in order to replace the procedure of pre-operative tibia CT scanning and segmentation.

%
%
\bibliographystyle{splncs04}
\bibliography{references}

\begin{thebibliography}{10}

\bibitem{gao2020primary}
Jiaxiang Gao, Dan Xing, Shengjie Dong, and Jianhao Lin.
\newblock The primary total knee arthroplasty: a global analysis.
\newblock {\em Journal of orthopaedic surgery and research}, 15:1--12, 2020.

\bibitem{pietrzak2019have}
Julien Pietrzak, Harold Common, Henri Migaud, Gilles Pasquier, Julien Girard,
  and Sophie Putman.
\newblock Have the frequency of and reasons for revision total knee
  arthroplasty changed since 2000? comparison of two cohorts from the same
  hospital: 255 cases (2013--2016) and 68 cases (1991--1998).
\newblock {\em Orthopaedics \& Traumatology: Surgery \& Research},
  105(4):639--645, 2019.

\bibitem{gromov2014optimal}
Kirill Gromov, Mounim Korchi, Morten~G Thomsen, Henrik Husted, and Anders
  Troelsen.
\newblock What is the optimal alignment of the tibial and femoral components in
  knee arthroplasty? an overview of the literature.
\newblock {\em Acta orthopaedica}, 85(5):480--487, 2014.

\bibitem{berend2004chetranjan}
Michael~E Berend et~al.
\newblock The chetranjan ranawat award: Tibial component failure mechanisms in
  total knee arthroplasty.
\newblock {\em Clinical Orthopaedics and Related Research (1976-2007)},
  428:26--34, 2004.

\bibitem{iorio2013accuracy}
R~Iorio et~al.
\newblock Accuracy of manual instrumentation of tibial cutting guide in total
  knee arthroplasty.
\newblock {\em Knee Surgery, Sports Traumatology, Arthroscopy},
  21(10):2296--2300, 2013.

\bibitem{patil2007improving}
Shantanu Patil, Darryl~D D'Lima, James~M Fait, and Clifford~W Colwell~Jr.
\newblock Improving tibial component coronal alignment during total knee
  arthroplasty with use of a tibial planing device.
\newblock {\em JBJS}, 89(2):381--387, 2007.

\bibitem{hetaimish2012meta}
Bandar~M Hetaimish, M~Moin Khan, Nicole Simunovic, Hatem~H Al-Harbi, Mohit
  Bhandari, and Paul~K Zalzal.
\newblock Meta-analysis of navigation vs conventional total knee arthroplasty.
\newblock {\em The Journal of arthroplasty}, 27(6):1177--1182, 2012.

\bibitem{mahfouz2003robust}
Mohamed~R Mahfouz, William~A Hoff, Richard~D Komistek, and Douglas~A Dennis.
\newblock A robust method for registration of three-dimensional knee implant
  models to two-dimensional fluoroscopy images.
\newblock {\em IEEE transactions on medical imaging}, 22(12):1561--1574, 2003.

\bibitem{kim2011novel}
Youngjun Kim, Kang-Il Kim, Jin hyeok Choi, and Kunwoo Lee.
\newblock Novel methods for 3d postoperative analysis of total knee
  arthroplasty using 2d--3d image registration.
\newblock {\em Clinical Biomechanics}, 26(4):384--391, 2011.

\bibitem{kobayashi2009automated}
Koichi Kobayashi, Makoto Sakamoto, Yuji Tanabe, Akihiro Ariumi, Takashi Sato,
  Go~Omori, and Yoshio Koga.
\newblock Automated image registration for assessing three-dimensional
  alignment of entire lower extremity and implant position using bi-plane
  radiography.
\newblock {\em Journal of Biomechanics}, 42(16):2818--2822, 2009.

\bibitem{baka20112d}
Nora Baka, Bart~L Kaptein, Marleen de~Bruijne, Theo van Walsum, JE~Giphart,
  Wiro~J Niessen, and Boudewijn~PF Lelieveldt.
\newblock 2d--3d shape reconstruction of the distal femur from stereo x-ray
  imaging using statistical shape models.
\newblock {\em Medical image analysis}, 15(6):840--850, 2011.

\bibitem{zheng20092d}
Guoyan Zheng, Sebastian Gollmer, Steffen Schumann, Xiao Dong, Thomas Feilkas,
  and Miguel A~Gonz{\'a}lez Ballester.
\newblock A 2d/3d correspondence building method for reconstruction of a
  patient-specific 3d bone surface model using point distribution models and
  calibrated x-ray images.
\newblock {\em Medical image analysis}, 13(6):883--899, 2009.

\bibitem{otake2011intraoperative}
Yoshito Otake, Mehran Armand, Robert~S Armiger, Michael~D Kutzer, Ehsan Basafa,
  Peter Kazanzides, and Russell~H Taylor.
\newblock Intraoperative image-based multiview 2d/3d registration for
  image-guided orthopaedic surgery: incorporation of fiducial-based c-arm
  tracking and gpu-acceleration.
\newblock {\em IEEE transactions on medical imaging}, 31(4):948--962, 2011.

\bibitem{markelj2012review}
Primoz Markelj, Dejan Toma{\v{z}}evi{\v{c}}, Bostjan Likar, and Franjo
  Pernu{\v{s}}.
\newblock A review of 3d/2d registration methods for image-guided
  interventions.
\newblock {\em Medical image analysis}, 16(3):642--661, 2012.

\bibitem{jones2018current}
Christopher~W Jones and Seth~A Jerabek.
\newblock Current role of computer navigation in total knee arthroplasty.
\newblock {\em The Journal of arthroplasty}, 33(7):1989--1993, 2018.

\bibitem{hampp2019robotic}
Emily~L Hampp et~al.
\newblock Robotic-arm assisted total knee arthroplasty demonstrated greater
  accuracy and precision to plan compared with manual techniques.
\newblock {\em The journal of knee surgery}, 32(03):239--250, 2019.

\bibitem{sassoon2015systematic}
Adam Sassoon, Denis Nam, Ryan Nunley, and Robert Barrack.
\newblock Systematic review of patient-specific instrumentation in total knee
  arthroplasty: new but not improved.
\newblock {\em Clinical Orthopaedics and Related Research{\textregistered}},
  473(1):151--158, 2015.

\bibitem{nam2013accelerometer}
Denis Nam, K~Durham Weeks, Keith~R Reinhardt, Danyal~H Nawabi, Michael~B Cross,
  and David~J Mayman.
\newblock Accelerometer-based, portable navigation vs imageless, large-console
  computer-assisted navigation in total knee arthroplasty: a comparison of
  radiographic results.
\newblock {\em The Journal of arthroplasty}, 28(2):255--261, 2013.

\bibitem{gao2019comparison}
Xiang Gao, Yu~Sun, Zhao-He Chen, Tian-Xu Dou, Qing-Wei Liang, and Xu~Li.
\newblock Comparison of the accelerometer-based navigation system with
  conventional instruments for total knee arthroplasty: a propensity
  score-matched analysis.
\newblock {\em Journal of orthopaedic surgery and research}, 14(1):1--9, 2019.

\bibitem{kirkpatrick1983optimization}
Scott Kirkpatrick, C~Daniel Gelatt~Jr, and Mario~P Vecchi.
\newblock Optimization by simulated annealing.
\newblock {\em science}, 220(4598):671--680, 1983.

\bibitem{nelder1965simplex}
John~A Nelder and Roger Mead.
\newblock A simplex method for function minimization.
\newblock {\em The computer journal}, 7(4):308--313, 1965.

\end{thebibliography}
\end{document}